\documentclass[10pt,twocolumn,letterpaper]{article}

\usepackage{cvpr}
\usepackage{times}
\usepackage{epsfig}
\usepackage{graphicx}
\usepackage{amsmath}
\usepackage{amssymb}
\usepackage{multirow}
\usepackage{graphicx}
\usepackage{subfigure}
\usepackage{mathrsfs}

\usepackage{color}
\usepackage{comment}
\usepackage{mdwlist}
\usepackage{authblk}
\newcommand{\newfootnote}[2]{\footnote{\label{#1}#2}}

\usepackage[pagebackref=true,breaklinks=true,letterpaper=true,colorlinks,bookmarks=false]{hyperref}

\cvprfinalcopy 

\def\cvprPaperID{160} 

\ifcvprfinal\fi
\begin{document}

\title{Oriented Response Networks}

\author[1]{Yanzhao Zhou}
\author[1]{Qixiang Ye}
\author[2]{Qiang Qiu}
\author[1]{Jianbin Jiao}

\affil[1]{University of Chinese Academy of Sciences}
\affil[2]{Duke University
\authorcr\small zhouyanzhao215@mails.ucas.ac.cn, \{qxye, jiaojb\}@ucas.ac.cn, qiang.qiu@duke.edu}

\renewcommand\Authands{ and }

\maketitle

\begin{abstract}
    Deep Convolution Neural Networks (DCNNs) are capable of learning unprecedentedly effective image representations.
    However, their ability in handling significant local and global image rotations remains limited.
    In this paper, we propose Active Rotating Filters (ARFs) that actively rotate during convolution and produce feature maps with location and orientation explicitly encoded.
    An ARF acts as a virtual filter bank containing the filter itself and its multiple unmaterialised rotated versions. During back-propagation, an ARF is collectively updated using errors from all its rotated versions.
    DCNNs using ARFs, referred to as Oriented Response Networks (ORNs), can produce within-class rotation-invariant deep features while maintaining inter-class discrimination for classification tasks.
    The oriented response produced by ORNs can also be used for image and object orientation estimation tasks.
    Over multiple state-of-the-art DCNN architectures, such as VGG, ResNet, and STN, we consistently observe that replacing regular filters with the proposed ARFs leads to significant reduction in the number of network parameters and improvement in classification performance.
    We report the best results on several commonly used benchmarks \newfootnote{fn:SourceCode}{Source code is publicly available at \href{http://yzhou.work/ORN/}{yzhou.work/ORN}}.
\end{abstract}

\section{Introduction}
    The problem of orientation information encoding has been extensively investigated in hand-crafted features, \eg, Gabor features \cite{Haley1995, Han2007}, HOG \cite{Dalal2005}, and SIFT \cite{Lowe1999}.
    In Deep Convolution Neural Networks (DCNNs), the inherent properties of convolution and pooling alleviate the effect of local transitions and warps; however, lacking the capability to handle large image rotation limits DCNN's performance in many visual tasks including object boundary detection \cite{Hallman2015, Maninis2016}, multi-oriented object detection \cite{Cheng2016}, and image classification \cite{Jaderberg2015, Laptev2016}.

    \begin{figure}
        \begin{center}
           \includegraphics[width=\linewidth]{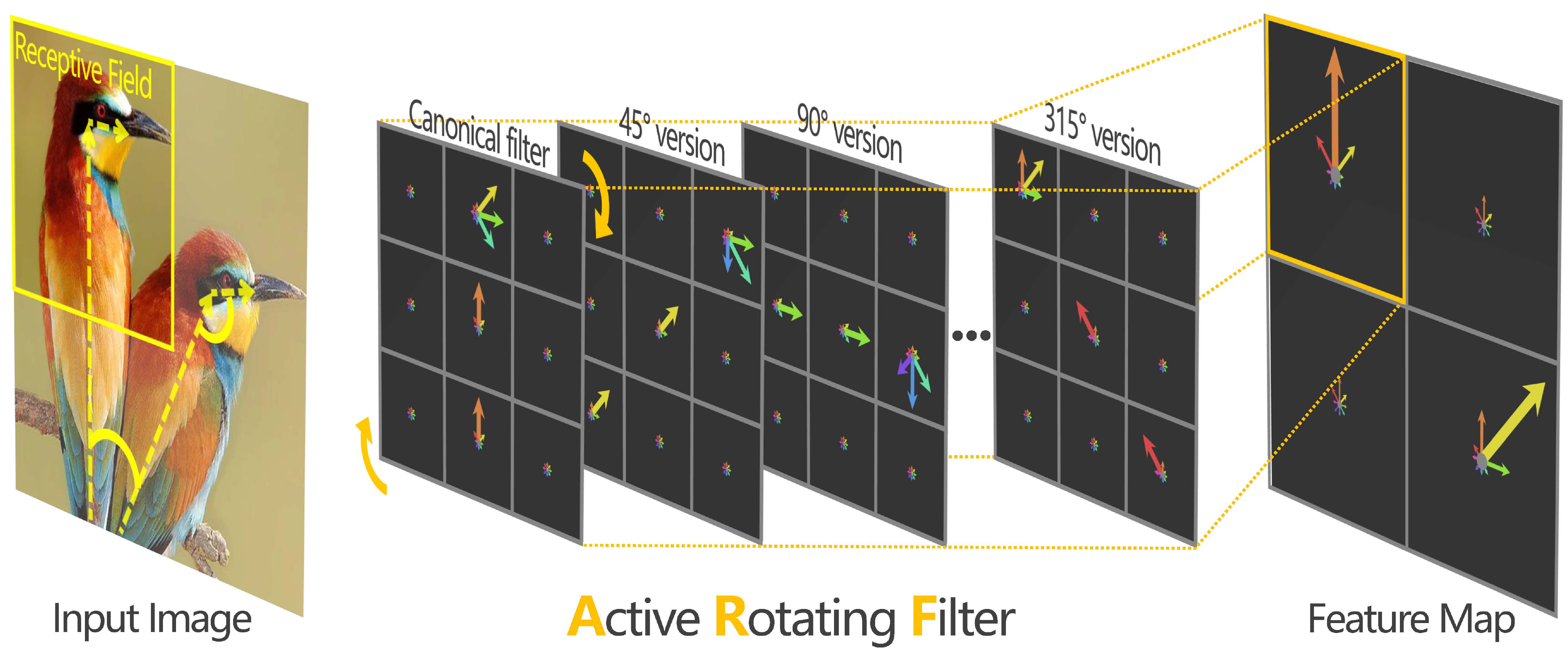}
        \end{center}
        \caption{
            An ARF is a filter of the size $W \times W\times N$, and viewed as N-directional points on a $W \times W$ grid.
            The form of the ARF enables it to effectively define relative rotations, \eg, the head rotation of a bird about its body.
            An ARF actively rotates during convolution; thus it acts as a virtual filter bank containing the canonical filter itself and its multiple unmaterialised rotated versions.
            In this example, the location and orientation of birds in different postures are captured by the ARF and explicitly encoded into a feature map.
        }
    \label{fig:cover}
    \vspace{-0.8em}
    \end{figure}

    Due to the lack of ability in fully understanding rotations, the most straightforward way for DCNN to decrease its loss is ``learning by rote''. The visualization of convolutional filters \cite{Erhan2009, Zeiler2014} indicates that different rotated versions of one identical image structure are often redundantly learned in low-level, middle-level, and relatively high-level filters, such as those in the VGG-16 model trained on ImageNet \cite{Deng2009}. When object parts rotate relatively to objects themselves, \eg, bird's head to its body, it requires learning multiple combinations of each orientation-distinct component with more convolutional filters. In such cases, the network could give up understanding the concept of the whole object and tend to use a discriminative part of it to make the final decisions \cite{Zhou2015}.
    The learning-by-rote strategy needs a larger number of parameters to generate orientation-redundant filters, significantly increasing both the training time and the risk of network over-fitting. Besides, the training data is not sufficiently utilized since the limited instances are implicitly split into subsets, which could increase the possibility of filter under-fitting. To alleviate such a problem, data augmentation, \eg, rotating each training sample into {multi-oriented versions}, is often used. Data augmentation improves the learning performance by extending the training set. However, it usually requires more network parameters and higher training cost.

    In this paper, we propose Active Rotating Filters (ARFs) and leverage Oriented Response Convolution (ORConv) to generate feature maps with orientation channels that explicitly encode the location and orientation information of discriminative patterns. Compared to conventional filters, ARFs have an extra dimension to define the arrangement of oriented structures. During the convolution, each ARF rotates and produces feature maps to capture the response of receptive fields from multiple orientations, as shown in Fig.~\ref{fig:cover}. The feature maps with orientation channels carry the oriented response along with the hierarchical network to produce high-level representations, endowing DCNNs the capability of capturing global/local rotations and the generalization ability for rotated samples never seen before.

    Instead of introducing extra functional modules or new network topologies, our method implements the prior knowledge of rotation to the most basic element of DCNNs, \ie, the convolution operator. Thus, it can be naturally fused with modern DCNN architectures, upgrading them to more expressive and compact Oriented Response Networks (ORNs). With the orientation information that ORNs produce, we can either apply SIFT-like feature alignment to achieve rotation invariance or perform image/object orientation estimation. The contributions of this paper are summarized as follows:

    \begin{itemize*}
        \item We specified Active Rotating Filters and Oriented Response Convolution, improved the most fundamental module of DCNN and endowed DCNN the capability of explicitly encoding hierarchical orientation information. We further applied such orientation information to rotation-invariant image classification and object orientation estimation.

        \item We upgraded successful DCNNs including VGG, ResNet, TI-Pooling and STN to ORNs, achieving state-of-the-art performance with significantly fewer network parameters on popular benchmarks.
    \end{itemize*}

\section{Related Works}
\subsection{Hand-crafted features.}
    Orientation information has been explicitly encoded in classical hand-crafted features including Weber's Law descriptor \cite{Chen2010}, Gabor features \cite{Haley1995,Han2007}, SIFT \cite{Lowe1999}, and LBP \cite{Ojala2002,Ahonen2006}. SIFT descriptor \cite{Lowe1999} and its modification with affine-local regions \cite{Lazebnik2004} find the dominant orientation of a feature point, according to which statistics of local gradient directions of image intensities are accumulated to give a summarizing description of local image structures. With dominant orientation based feature alignment, SIFT achieves invariance to rotation and robustness to moderate perspective transforms \cite{Bicego2006,Goesele2007}. Starting from the gray values of a circularly symmetric neighbor set of pixels in a local neighborhood, LBP derives an operator that is by definition invariant against any monotonic transformation of the gray scale \cite{Ojala2002,Ahonen2006}. Rotation invariance is achieved by minimizing the LBP code value using the bit cyclic shift. Other representative descriptors including CF-HOG \cite{Skibbe2012} that uses orientation alignment and RI-HOG \cite{Liu2014} that leverages radial gradient transform to be rotation invariant.

    \begin{figure*}
        \begin{center}
            \includegraphics[width=0.8\linewidth]{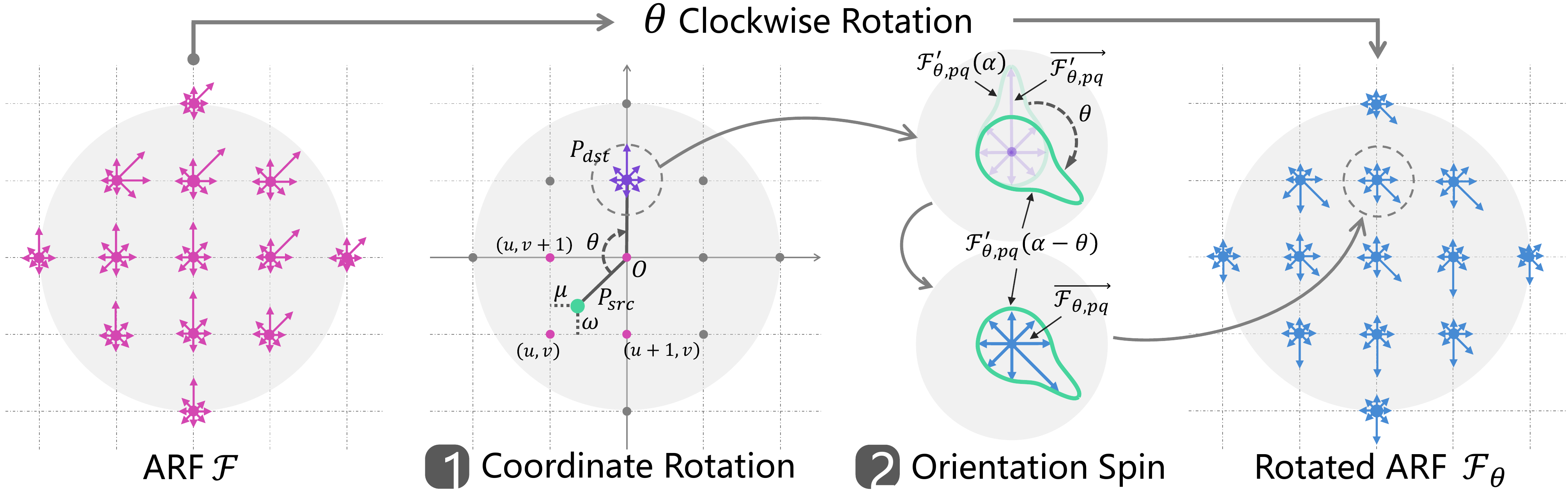}
        \end{center}
        \caption{
            An ARF $\mathcal{F}$ is clockwise rotated by $\theta$ to yield its rotated variant $\mathcal{F}_{\theta}$ in two steps: coordinate rotation and orientation spin.
        }
    \label{fig:RotARF} 
    \vspace{-0.6em}
    \end{figure*}

\subsection{Deep Convolutional Neural Networks.}
    Deep Convolution Neural Networks have the capability of processing transforms including moderate transitions, scale changes, and small rotations. Such capability is endowed with the inherent properties of convolutional operations, redundant convolutional filters, and hierarchical spatial pooling \cite{Scherer2010,Jaderberg2015}. More general pooling operations \cite{Lee2016} permit to consider invariance to local deformation that however does not correspond to specific prior knowledge.

    \textbf{Data augmentation.}
    Given rich, and often redundant, convolutional filters, data augmentation can be used to achieve local/global transform invariance \cite{VanDyk2012}. Despite the effectiveness of data augmentation, the main drawback lies in that learning all the possible transformations of augmented data usually requires more network parameters, which significantly increases the training cost and the risk of over-fitting. Most recent TI-Pooling \cite{Laptev2016} alleviates the drawbacks by using parallel network architectures for the considered transform set and applying the transform invariant pooling operator on their outputs before the top layer. The essence of TI-Pooling comprises multi-instance learning and weight sharing which help to find the most optimal canonical instance of the input images for training, as well as reducing the redundancy in learned networks. Nevertheless, with built-in data augmentation, TI-Pooling requires significantly more training and testing cost than a standard DCNN.

    \textbf{Spatial Transform Network.}
    Representatively, the spatial transformer network (STN) \cite{Jaderberg2015} introduces an additional network module that can manipulate the feature maps according to the transform matrix estimated with a localisation sub-CNN. STN contributes a general framework for spatial transform, but the problem about how to precisely estimate the complex transform parameters by CNN remains not being well-solved \cite{Goodfellow2014, Radford2015}.
    In \cite{Kivinen2011, Schmidt2012}, the Convolutional Restricted Boltzmann Machine (C-RBM) induces transformation-aware filters, \ie, it yields filters that have a notion with which specific image transformation they are used.
    From the view of group theory, Cohen \etal~\cite{Cohen2016} justified that the spatial transform of images could be reflected in both feature maps and filters, providing a theoretical foundation for our work.
    Most recent works \cite{Wu2015,Marcos2016} have tried rotating conventional filters to perform rotation-invariant texture and image classification; however, without upgrading conventional filters to multi-oriented filters with orientation channels, their capability about capturing hierarchical and fine-detailed orientation information remains limited.


\section{Oriented Response Networks}
    Oriented Response Networks (ORNs) are deep convolutional neural networks using Active Rotating Filters (ARFs).
    An ARF is a filter that actively rotates during convolution to produce a feature map with multiple orientation channels. Thus, an ARF acts as a virtual filter bank with only one filter being materialized and learned.
    With ARFs, ORNs require significantly fewer network parameters with negligible computation overhead and enable explicitly hierarchical orientation information encoding.

    In what follows, we address three problems in adopting ARFs in DCNN.
    First, we construct a two-step technique to efficiently rotate an ARF based on the circular shift property of Fourier Transform.
    Second, we describe convolutions that use ARFs to produce feature maps with location and orientation explicitly encoded.
    Third, we show how all rotated versions of an ARF contribute to its learning during the back-propagation update stage.

\subsection{Active Rotating Filters}
\label{sec:ARF}
    An Active Rotating Filter (ARF) is a filter of the size $W \times W \times N$ that actively rotates $N-1$ times during convolution to produce a feature map of $N$ orientation channels, Fig.~\ref{fig:RotARF}.
    Therefore, an ARF $\mathcal{F}$ can be virtually viewed as a bank of $N$ filters ($N \times W \times W \times N$), where only the canonical filter $\mathcal{F}$ itself is materialized and to be learned, and the remaining $N-1$ filters are its unmaterialized copies. The $n$-th filter in such {a} filter bank, $n \in [1, N-1]$, is obtained by clockwise rotating $\mathcal{F}$ by $\frac{2\pi n}{N}$.

    An ARF contains $N$ orientation channels and is viewed as $N$-directional points on a $W \times W$ grid. Each element in an ARF $\mathcal{F}$ can be accessed with $\overrightarrow{\mathcal{F}_{ij}}^{(n)}$ where $0 \leq |i|,|j| \leq \frac{W-1}{2}, 0 \leq n \leq N-1, i,j,n \in \mathbb{N}$. An ARF $\mathcal{F}$ is clockwise rotated by $\theta$ to yield its rotated variant $\mathcal{F}_{\theta}$ through the following two steps, coordinate rotation and orientation spin.

    \textbf{Coordinate Rotation}.
        An ARF rotates around the origin $O$, Fig.~\ref{fig:RotARF}, and the point at $(p,q)$ in $\mathcal{F}_{\theta}$ is calculated from four neighbors around $(p',q')$ in $\mathcal{F}$,
        $\left(\begin{smallmatrix}p'&q'\end{smallmatrix}\right)=\left(\begin{smallmatrix}p&q\end{smallmatrix}\right)
        \left(\begin{smallmatrix}cos(\theta)&sin(\theta)\\-sin(\theta)&cos(\theta)\end{smallmatrix}\right)$,
        using bilinear interpolation
        \begin{equation}
            \begin{aligned}
                \overrightarrow{\mathcal{F'}_{\theta, pq}} &= (1-\mu)(1-\omega)\overrightarrow{\mathcal{F}_{uv}} +
                (1-\mu)\omega\overrightarrow{\mathcal{F}_{u,v+1}} \\
                &+ \mu(1-\omega)\overrightarrow{\mathcal{F}_{u+1,v}} + \mu\omega\overrightarrow{\mathcal{F}_{u+1,v+1}},
            \end{aligned}
        \label{eq:Interpolation}
        \end{equation}
        where  $u=\lfloor p' \rfloor, v=\lfloor q' \rfloor, \mu = p'-u, \omega = q'-v$.
        Note that points outside the inscribed circle are padded with 0.

    \textbf{Orientation Spin}.
        As discussed, an ARF can be viewed as $N$-directional points on a grid. Each $N$-directional point $\overrightarrow{\mathcal{F'}_{\theta, pq}}$ is the $N$-points uniform sampling of a desired oriented response $\mathcal{F'}_{\theta, pq}(\alpha)$, which is a continuous periodic function of angle $\alpha$ with period $2\pi$.
        After the coordinates rotation, it still requires a clockwise spin by $\theta$ to yield $\overrightarrow{\mathcal{F}_{\theta, pq}}$, which is, in fact, the quantization of $\mathcal{F'}_{\theta, pq}(\alpha-\theta)$, {Fig.~\ref{fig:RotARF}}. Therefore, such spin procedure can be efficiently tackled in Fourier domain by using the circular shift property of Discrete Fourier Transforms (DFT),
        \begin{equation}
            \begin{aligned}
                X(k) &\equiv \mathbf{DFT}\{\overrightarrow{\mathcal{F'}_{\theta, pq}}^{(n)}\} \\
                    &= \sum_{n=0}^{N-1} \overrightarrow{\mathcal{F'}_{\theta, pq}}^{(n)} e^{-j k \frac{2\pi n}{N}},
                    {\scriptstyle k=0,1,...,N-1}, \\
            \end{aligned}
        \label{eq:Spin}
        \end{equation}

        \begin{equation}
            \begin{aligned}
            \overrightarrow{\mathcal{F}_{\theta, pq}}^{(n)} &\equiv \mathbf{IDFT}\{X(k)e^{-j k \theta}\} \\
                    &= \frac{1}{N}\sum_{k=0}^{N-1}X(k)e^{j k (\frac{2\pi n}{N} - \theta) }, {\scriptstyle n=0,1,...,N-1}.
            \end{aligned}
        \label{eq:Spin2}
        \end{equation}

        To smoothly process all rotation angles, ARFs require a considerable amount of orientation channels. In practice, thanks to the orientation `interpolation' by multi-layer pooling operations, we can use a limited amount of orientations to guarantee the accuracy. The successful practice of DCNNs, \eg, VGG \cite{Simonyan2014} and ResNet \cite{He2015, He2016}, shows that the stacks of multiple small filters are more expressive and parameters-efficient than large filters. Moreover, when using the combination of small filters and a limited number of orientation channels, the computational complexity of rotating ARF can be further reduced, since both the coordinate rotation and the orientation spin can be calculated by the circular shift operator and implemented via high-efficient memory mapping under reasonable approximations. Take a $3\times3\times8$ ARF $\hat{\mathcal{F}}$ as an example, calculations of its $\theta$ clockwise rotated version $\hat{\mathcal{F}_{\theta}}$ are formulated as
        \begin{equation}
            \begin{aligned}
                \overrightarrow{\hat{\mathcal{F}'}_{\theta, \langle i \rangle}} &= \overrightarrow{\hat{\mathcal{F}'}_{\langle (i-k)\
                \mathbf{mod}\ N \rangle}}, {\scriptstyle i \in \mathcal{I}}, \\
                \overrightarrow{\hat{\mathcal{F}}_{\theta}}^{(n)} &= \overrightarrow{\hat{\mathcal{F}'}_{\theta}}^{((n-k)\
                \mathbf{mod}\ N)}, {\scriptstyle n=0,1,...,N-1},
            \end{aligned}
            \label{eq:SmallARF}
        \end{equation}
        where $\forall{k} \in \mathbb{N}, \theta = k \frac{2\pi}{N}, N = 8$ and $\mathcal{I} = \left(\begin{smallmatrix}7&0&1\\6&
        &2\\5&4&3\end{smallmatrix}\right)$ is a mapping table that defines the index of each surrounding element, which means
        $\overrightarrow{\hat{\mathcal{F}}_{\langle 0 \rangle}} \equiv \overrightarrow{\hat{\mathcal{F}}_{0,1}}$,
        $\overrightarrow{\hat{\mathcal{F}}_{\langle 1 \rangle}} \equiv \overrightarrow{\hat{\mathcal{F}}_{1,1}}$,
        $\overrightarrow{\hat{\mathcal{F}}_{\langle 2 \rangle}} \equiv \overrightarrow{\hat{\mathcal{F}}_{1,0}}$,
        $\overrightarrow{\hat{\mathcal{F}}_{\langle 3 \rangle}} \equiv \overrightarrow{\hat{\mathcal{F}}_{1,-1}}$ and so on.

        Given the above, we use $1\times1$ and $3\times3$ ARFs with $4$ and $8$ orientation channels in most experiments.

        \begin{figure}[t]
            \begin{center}
                \includegraphics[width=1.0\linewidth]{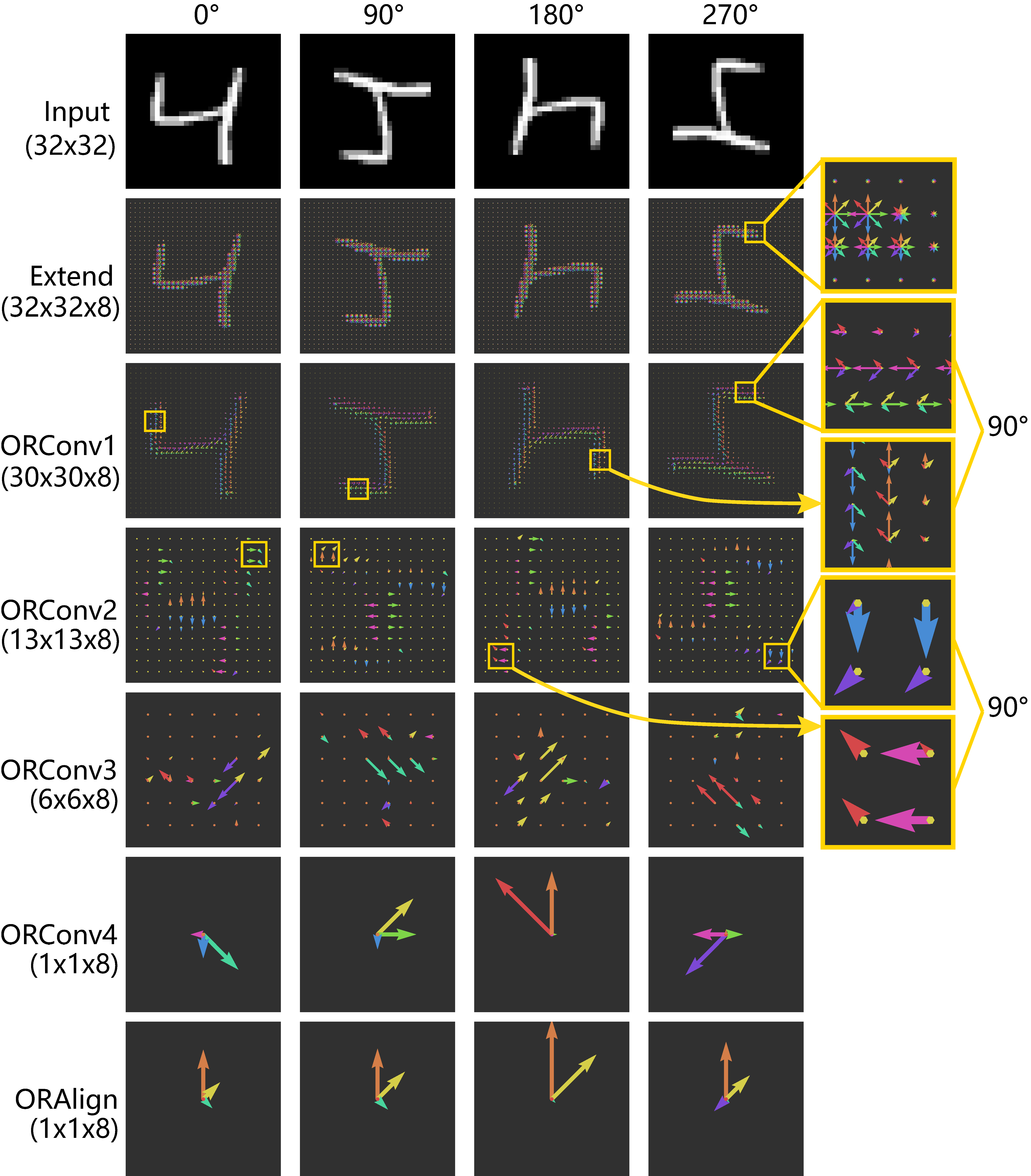}
            \end{center}
            \caption{
                Example feature maps produced by one ARF at each layer of an ORN trained on the rotated MNIST dataset, with digit `4' in different rotations as the inputs (one network layer per row, one input per column).
                The right-most column magnifies sample regions in feature maps.
                It clearly shows that a feature map explicitly encodes position and orientation.
                At the second layer, an image is extended to an omnidirectional map to fit ORConv.
                At the second-to-last (ORConv4) layer, deep features are observed in similar values but in different orientations, which demonstrates that orientation information is extracted by ORNs.
                The last (ORAlign) layer performs SIFT-like alignment to enable rotation-invariance
                (Best viewed zooming on screen).
            }
        \label{fig:OFTM}
        \end{figure}

\subsection{Oriented Response Convolution}
    An ARF actively rotates $N-1$ times during convolution to produce a feature map of $N$ orientation channels, and such feature map explicitly encodes both location and orientation information.
    As an ARF is defined as the size $W \times W \times N$, both an ARF $\mathcal{F}$ and an $N$-channel feature map $\mathcal{M}$ can be viewed as $N$-directional points on a grid.
    With ARF, we define the Oriented Response Convolution over $\mathcal{F}$ and $\mathcal{M}$, denoted as $\tilde{\mathcal{M}} = \mathbf{ORConv}(\mathcal{F},\mathcal{M})$.
    The output feature map $\tilde{\mathcal{M}}$ consists of $N$ orientation channels and the $k$-th channel is computed as
    \begin{equation}
       \tilde{\mathcal{M}}^{(k)}  = \sum_{n=0}^{N-1}\mathcal{F}_{\theta_k}^{(n)} \ast \mathcal{M}^{(n)},
       \theta_k = k\frac{2\pi}{N}, {\scriptstyle k=0,...,N-1},
    \label{eq:ORConv}
    \end{equation}
    where $\mathcal{F}_{\theta_k}$ is the clockwise $\theta_k$-rotated version of $\mathcal{F}$, $\mathcal{F}_{\theta_k}^{(n)}$ and $\mathcal{M}^{(n)}$ are the $n$-th orientation channel of $\mathcal{F}_{\theta_k}$ and $\mathcal{M}$ respectively.

    According to (\ref{eq:ORConv}), the $k$-th orientation channel of the output feature map $\tilde{\mathcal{M}}$ is generated by $\theta_k$ rotated versions of the materialised ARF. It means that in each oriented response convolution, the ARF proactively captures image response in multiple directions and explicitly encodes its location and orientation into a single feature map with multiple orientation channels, visualized in Fig.~\ref{fig:OFTM}. (\ref{eq:ORConv}) also demonstrates that each orientation channel of the ARF contributes to the final convolutional response respectively, endowing ORNs the capability of capturing richer and more fine-detailed patterns than a regular CNN.

\subsection{Updating Filters}
\begin{figure}[t]
        \begin{center}
            \includegraphics[width=1\linewidth]{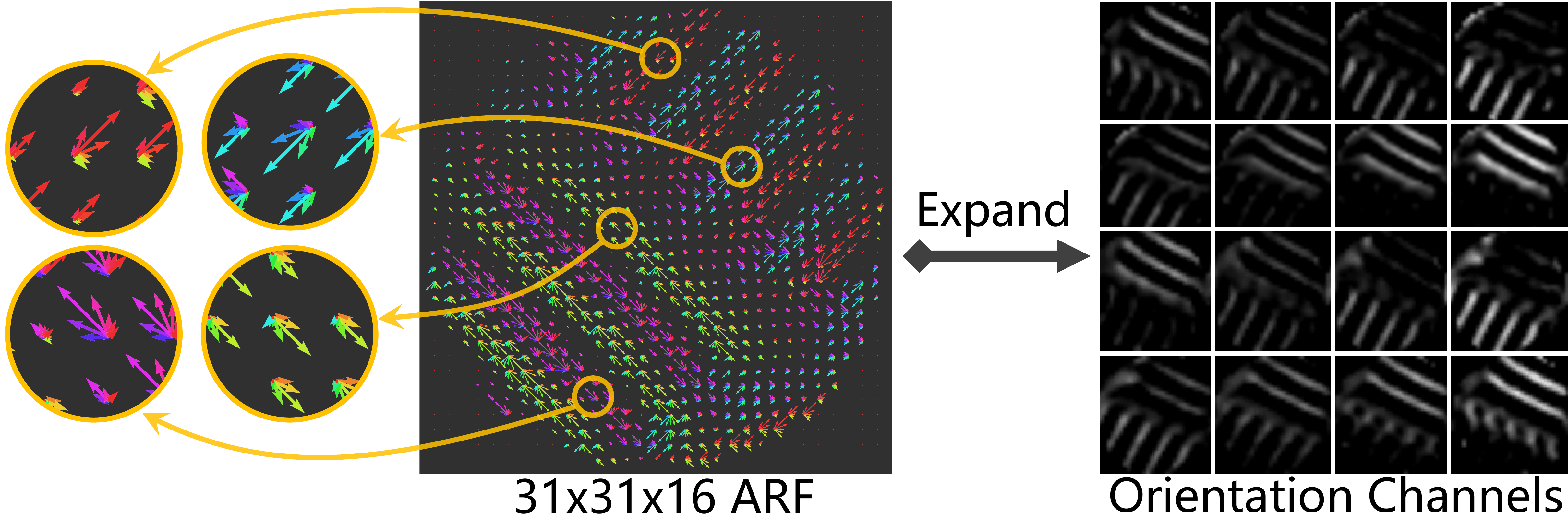}
        \end{center}
        \caption{
            A $31\times31\times16$ ARF learned from a texture dataset.
            It is shown in the N-directional points form (left) and further visualized as one orientation channel per image (right).
            The ARF clearly defines a texture pattern through a combination of multi-oriented edges
            (Best viewed zooming on screen).
        }
    \label{fig:ARF-Channels}
    \vspace{-0.8em}
    \end{figure}

    During the back-propagation, error signals $\delta^{(k)}$ of all rotated versions of the ARF are aligned to $\delta^{(k)}_{-\theta_k}$ using (\ref{eq:Interpolation}) and (\ref{eq:Spin}), and aggregated to update the materialised ARF,
    \begin{equation}
        \begin{aligned}
            \delta^{(k)} &= \frac{\partial L}{\partial \mathcal{F}_{\theta_k}}, \theta_k = k\frac{2\pi}{N}, {\scriptstyle k=0,1,...,N-1},  \\
            \mathcal{F} &\leftarrow \mathcal{F} - \eta\sum_0^{N-1} \delta^{(k)}_{-\theta_k},
        \end{aligned}
    \label{eq:Backprop}
    \end{equation}
    where $L$ stands for training loss and $\eta$ for learning rate.
    An ARF acts as a virtual filter bank containing the materialized canonical filter itself and unmaterialised rotated versions. According to (\ref{eq:Backprop}), the back-propagation collectively updates the materialised filter only, so that training errors of appearance-like but orientation-distinct samples are aggregated. In low-level layers, such collective updating contributes more significantly, as in a single image there exist many appearance-like but orientation-distinct patches that can be exploited.
    The collective updating also helps when only limited training samples are given.
    One example of a collectively updated ARF is shown in Fig.~\ref{fig:ARF-Channels}.

\subsection{Rotation Invariant Feature Encoding}
\label{sec:rot-inv-encoding}
    Feature maps in ORNs are not rotation-invariant as orientation information are encoded instead of being discarded.
    When within-class rotation-invariance is required, we introduce two strategies, ORAlign and ORPooling, at the top layer of ORNs.
    For simplicity, we choose a DCNN architecture, where the size of a feature map gradually shrinks to $1 \times 1 \times N$. $N$ is the number of orientation channels. Each {feature map} of the last ORConv layer has a receptive field of image size and stands for the oriented response of high-level representative patterns.

    The first strategy is the ORAlign. Without loss of generality, let us denote the $i$-th {feature map} of the last ORConv layer as $\overrightarrow{\hat{\mathcal{M}}\{i\}}$ and each oriented response in it as $\overrightarrow{\hat{\mathcal{M}}\{i\}}^{(n)}, 0 \leq n \leq N-1$.  $\overrightarrow{\hat{\mathcal{M}}\{i\}}$ is an $N$ dimension tensor records the response from different directions, with which we perform SIFT-like alignment to achieve rotation robustness. This is done by first calculating the dominant orientation (the orientation with the strongest response) as $D = {\underset{d}{\mathrm{argmax}}}\overrightarrow{\hat{\mathcal{M}}\{i\}}^{(d)}$ and spin the feature by $-D\frac{2\pi}{N}$, Fig.~\ref{fig:OFTM}.
    The second strategy is the ORPooling, which is done via simply pooling a $\overrightarrow{\hat{\mathcal{M}}\{i\}}$ to a scalar $\max(\overrightarrow{\hat{\mathcal{M}}\{i\}}^{(j)}),0<j<N-1$.
    This strategy reduces the feature dimension but loses feature arrangement information.

\section{Experiments}
    ORNs are evaluated on three benchmarks. In Sec.~\ref{sec:MNIST-rot}, experiments on the MNIST dataset \cite{Liu2003} and its $[0,2\pi]$ randomly rotated versions are conducted, showing the advantage of ORNs through encoding rotation-invariant features, and reducing network parameters. ORNs are further tested on a small sample set of $[0,2\pi]$ rotated MNIST \cite{Larochelle2007} to validate its generalization ability on rotation.
    In Sec.~\ref{sec:MNIST-estimate}, on a weakly-supervised orientation estimate task, the vast potential of directly taking advantage of the orientation information extracted by ORNs is demonstrated.
    In Sec.~\ref{sec:CIFAR}, we upgrade the VGG \cite{Simonyan2014}, ResNet \cite{He2015}, and the WideResNet \cite{Zagoruyko2016} to ORNs, and {train} them on CIFAR10 and CIFAR100 \cite{Krizhevsky2009}, showing the state-of-the-art performance on the natural image classification task.

\subsection{Rotation Invariance}
\label{sec:MNIST-rot}
    \begin{table*}[t!]
        \begin{center}
            \begin{tabular}{|l|cccccc|}
                \hline
                Method            & time(s)      & params(\%)    & original(\%)  & rot(\%)       & rot+(\%)      & original
                $\rightarrow$ rot(\%) \\ \hline\hline
                Baseline CNN      & 16.4         & 100.00        & 0.73          & 2.82          & 2.19          & 56.28
                \\
                STN(affine)\cite{Jaderberg2015}    & 18.5         & 100.40        & 0.61          & 2.52          & 1.82          &
                56.44                        \\
                STN(rotation)\cite{Jaderberg2015}  & 18.7         & 100.39        & 0.66          & 2.88          & 1.93          &
                55.59                        \\
                TIPooling(x8)\cite{Laptev2016}       & 126.7        & 100.00        & $0.97^{\dagger}$          & not permitted & 1.26
                & not permitted                \\ \hline
                ORN-4(None)    & 7.9          & 15.91         & 0.63          & 1.88          & 1.55          & 59.67
                \\
                ORN-4(ORPooling) & 8            & 7.95          & 0.59          & 1.84          & 1.33          & 27.74
                \\
                ORN-4(ORAlign)   & 8.1          & 15.91         & \textbf{0.57} & 1.69          & 1.34          & 27.92
                \\
                ORN-8(None)    & 17.5         & 31.41         & 0.79          & 1.57          & 1.33          & 58.98
                \\
                ORN-8(ORPooling) & 17.9         & 12.87         & 0.66          & \textbf{1.37} & 1.21          & 16.67
                \\
                ORN-8(ORAlign)   & 17.8         & 31.41         & 0.59          & 1.42          & \textbf{1.12}& \textbf{16.21} \\
                \hline
            \end{tabular}
        \end{center}
        \caption{
            Results on the MNIST variants. The second column describes the average training time of an epoch on the \textit{original} training set (with a NVIDIA Tesla K80 GPU).
            The third column describes the percentage of parameters of each model about the baseline CNN. The fourth to sixth columns describe the error rates on the \textit{original}, the \textit{rot}, and the \textit{rot+} datasets. The last column describes the error rates achieved on the \textit{rot} testing set (with random rotation) by models trained on the \textit{original} training set (without rotation). TIPooling requires augmented data; thus some experiments are not permitted. The error rate of TIPooling on the original MNIST dataset is under augmentation, with the superscript $\dagger$ to show its difference with others.
        }
    \label{tab:MNIST-rot}
    \vspace{-0.4em}
    \end{table*}

    \begin{figure}[t]
        \begin{center}
            \includegraphics[width=0.8\linewidth]{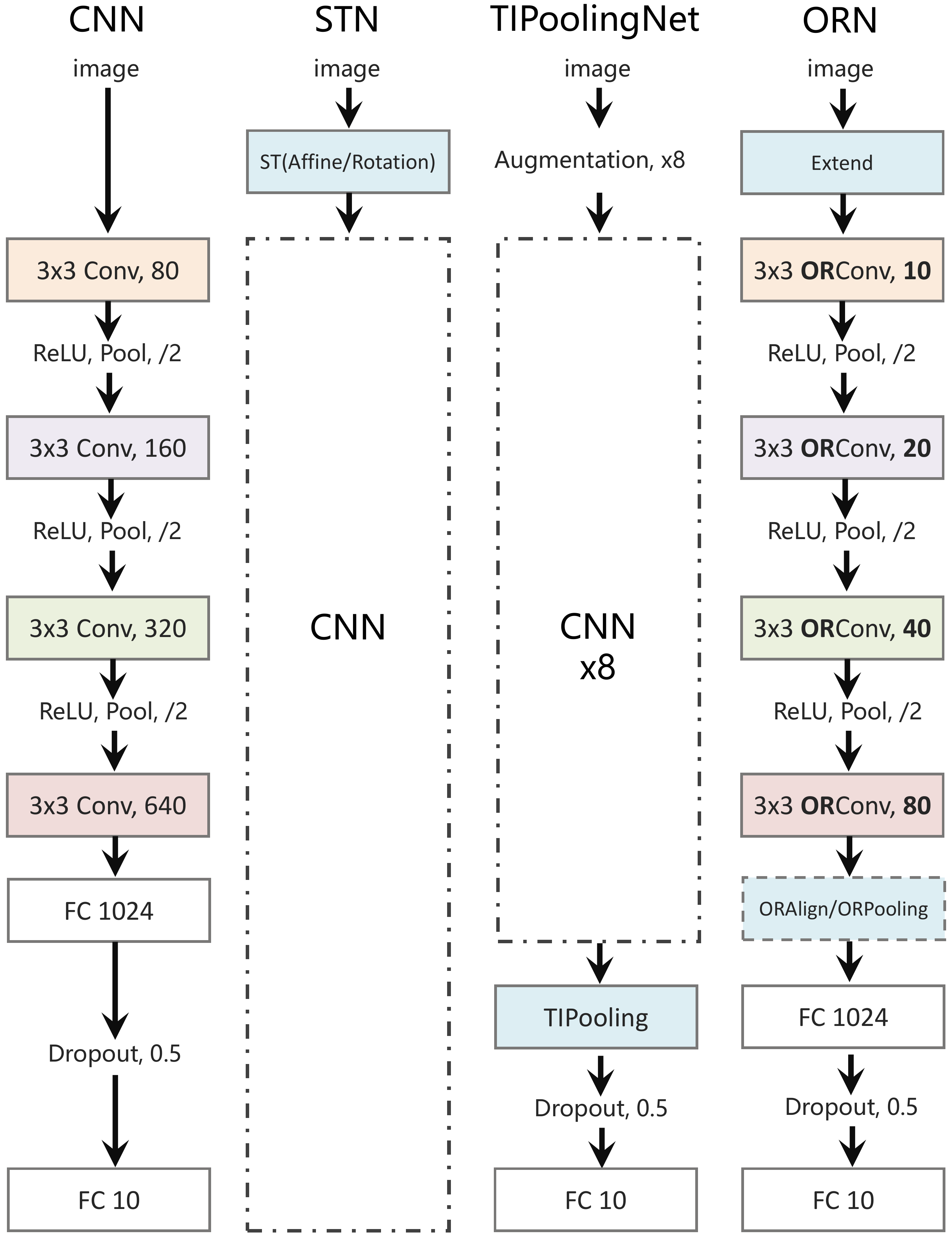}
        \end{center}
        \caption{Comparison of network topologies.}
    \label{fig:Topologies}
    \vspace{-1em}
    \end{figure}

    \textbf{Rotated MNIST.}
    We randomly rotate each sample in the MNIST dataset \cite{Liu2003} between $[0, 2\pi]$ to yield \textit{MNIST-rot}. To assess the effect of data augmentation on different models, we further rotate each sample in the \textit{MNIST-rot} training set to eight directions with 45-degree intervals, which means that the training set is augmented eightfold. The augmented data set is identified as \textit{MNIST-rot}+.

    We set up a baseline CNN with four convolutional layers and multiple 3x3 filters, Fig.~\ref{fig:Topologies}.
    With the baseline CNN, we generate different ORNs, as well as configuring the STNs \cite{Jaderberg2015} and the TI-Pooling network \cite{Laptev2016} for comparison.
    STNs are created by inserting a Spatial Transformer with affine or rotation transform to the entry of the baseline CNN.
    TIPooling network is constructed by duplicating the baseline CNN eight times to capture different augmented rotated versions of inputs, and a transform-invariant pooling layer before the output layer. ORNs are built by upgrading each convolution layer in the baseline CNN to Oriented Response Convolution layer using Active Rotating Filters (ARFs) with 4 or 8 orientation channels. Considering that ARFs are more expressive than conventional filters, the number of ARFs in each layer is decreased to one-eighth of those in the baseline. Corresponding to the strategies proposed in Sec.~\ref{sec:rot-inv-encoding}, we use ORAlign, ORPooling or none of them to encode rotation-invariant features. The network topologies are shown in Fig.~\ref{fig:Topologies}.

    In network training, we use the same hyper-parameters as TI-Pooling \cite{Laptev2016}, \ie, 200 training epochs using the turning-free convergent adadelta algorithm \cite{Zeiler2012}, 128 batch size, and 0.5 dropout rate for the fully-connected layer. For each dataset, we randomly selected 10,000 samples from the training set for validation and the remaining 50,000 samples for training. The best model selected by 5-fold cross-validation is then applied to the test set, and the final results are {presented} in Tab.~\ref{tab:MNIST-rot}.

    \begin{figure}
        \centering
        \includegraphics[width=0.75\linewidth]{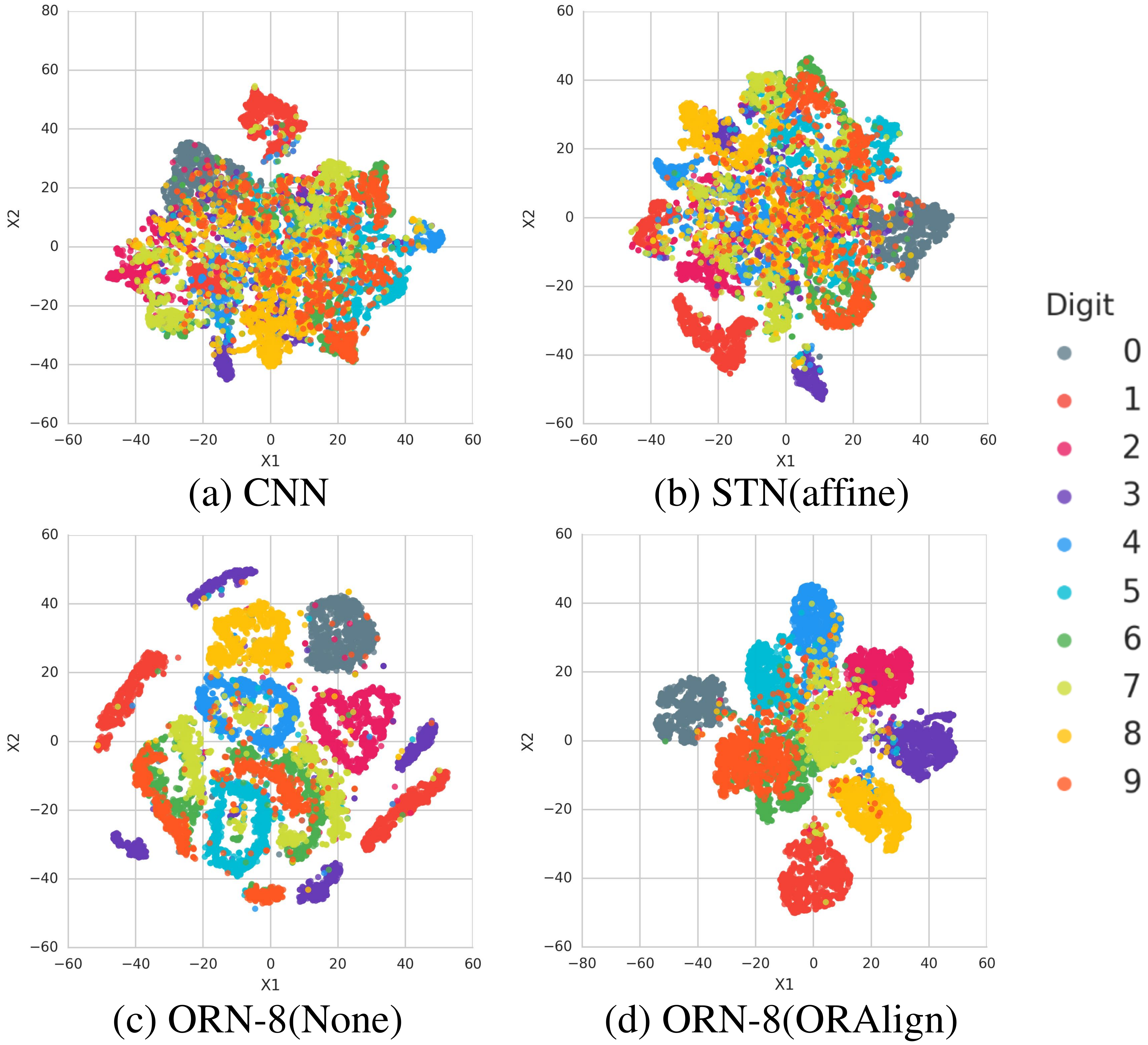}
        \caption{Visualization of features in cross-generalization evaluation, corresponding to the last column of Tab.~\ref{tab:MNIST-rot}.}
    \label{fig:FeatureVis}
    \vspace{-0.8em}
    \end{figure}

    The second column of Tab.~\ref{tab:MNIST-rot} shows that ORN keeps high training efficiency. The ORN-4 (4 orientation channels) uses only 50\% training time while ORN-8 uses similar training time with the baseline CNN. In {contrast},
    TIPooling increases the time by about eight times as each sample is augmented to 8 orientations. From the third to the last column of Tab.~\ref{tab:MNIST-rot}, it can be seen that ORNs can use significantly fewer network parameters (7.95\%-31.4\%) to consistently improve the performance. Even on the original dataset without sample rotations, it achieves 22\% error rate decrease (0.57\% vs 0.73\%), as the digit curvatures are well modeled by ORN. Compared with the data augmentation strategy (baseline CNN on \textit{rot+}), ORN (on \textit{rot}) not only reduces network parameters and training cost but also achieves significant lower error rate (1.37\% vs 2.19\%).

    Tab.~\ref{tab:MNIST-rot} also shows that different rotation-invariant encoding strategies have different advantages. ORPooling can further compress the feature dimension and network parameters, while ORAlign retains the complete feature structure thus achieves higher performance. Even without rotation-invariant encoding, ORNs outperforms the baseline on the \textit{rot} and \textit{rot+}, because ARFs can explicitly capture the response in different directions so that a pattern and its rotated versions can be encoded in the same feature map with orientation channels, Fig.~\ref{fig:OFTM}.
    It also can be seen in Fig.~\hyperref[fig:FeatureVis]{\ref*{fig:FeatureVis}(c)} that the t-SNE \cite{Maaten2008} 2D mapping of features produced by ORN-8(None) constitutes clear clusters.

    In Tab.~\ref{tab:MNIST-rot}, the state-of-the-art spatial transform network, STN, has minor improvement on the \textit{rot} while slightly increasing the number of parameters. The visualization of calibrated images shows that it often outputs wrong transform parameters. This validates our previous viewpoint: the conventional CNN used in STN lacks the capability to precisely estimate rotation parameters. In Sec.~\ref{sec:MNIST-estimate}, we will show that ORN can better solve such a problem.

    The last column of Tab.~\ref{tab:MNIST-rot} presents the results of cross-generalization evaluation that trains models on the \textit{MNIST-original} and tests them on the \textit{MNIST-rot}. ORNs show impressing performance with 71\% improvement over the baseline. Fig.~\hyperref[fig:FeatureVis]{\ref*{fig:FeatureVis}(d)} shows that ORN-8(ORAlign) produces much clearer feature distribution in manifold than other networks.

\begin{figure}
        \centering
        \includegraphics[width=1\linewidth]{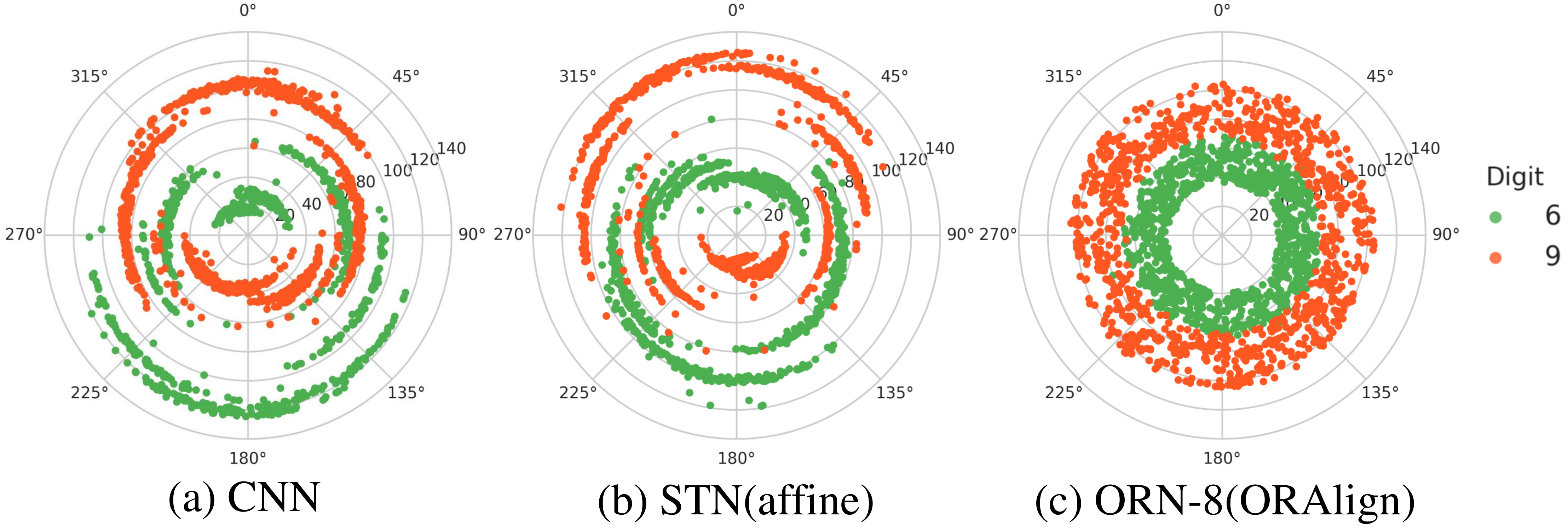}
        \caption{
            Visualization of features encoding of digit class `6' and `9' from \textit{MNIST-rot}. Each point ($r, \theta$) corresponds to a sample where radius \textit{r} is the 1-D tSNE feature mapping, and $\theta$ is the angle of the sample.
            ORN-8(ORAlign) produces within-class rotation-invariant deep features while maintaining inter-class discrimination. (Best viewed in color.)
        }
    \label{fig:RotFeatureVis}
    \vspace{-0.8em}
    \end{figure}

    An interesting experiment comes from the digit class `6' and `9'. It can be seen in Fig.~\ref{fig:RotFeatureVis} that both CNN and STN have large within-class differences as the same digit with different angles produce various radii. Moreover, features generated by CNN and STN have apparently $180^o$ symmetrical distribution, which means that they can barely tell the difference between upside-down 6 and 9.
    In contrast, ORN-8(ORAlign) generates within-class rotation-invariant deep features, while maintaining inter-class discrimination.

    \begin{table}
        \begin{center}
            \begin{tabular}{|l|c|}
                \hline
                Method                          & Error(\%)     \\ \hline\hline
                ScatNet-2 \cite{Bruna2013}                    & 7.48          \\
                PCANet-2 \cite{Chan2015}                     & 7.37          \\
                TIRBM \cite{Sohn2012}                        & 4.2           \\ \hline
                CNN                             & 4.34          \\
                ORN-8(ORAlign)               & \textbf{2.25} \\ \hline
                TIPooling(with augmentation) \cite{Laptev2016}   & 1.93          \\
                OR-TIPooling(with augmentation) & \textbf{1.54} \\ \hline
            \end{tabular}
        \end{center}
        \caption{Classification error rates on the \textit{MNIST-rot-12k}.}
    \label{tab:MNIST-rot-small}
    \end{table}

    \textbf{Rotated Small Sample Set.} A smaller dataset can better test the generalization capability of a learning model. We consider the \textit{MNIST-rot-12k} dataset \cite{Larochelle2007} which contains 12,000 training samples and 50,000 test samples from the \textit{MNIST-rot} dataset. Among them, 2000 training samples are used as the validation set and the remaining 10,000 samples as the training set.

    In the dataset, we test the ORN-8 model that uses 8-orientation ARFs and an ORAlign operator. We also test the OR-TIPooling network, which is constructed by upgrading its parallel CNNs to ORN-8(None)s. The reason why we do not use ORAlign or ORPooling is that TIPooling itself has the invariant encoding operator. Tab.~\ref{tab:MNIST-rot-small} shows that ORN can decrease the state-of-the-art error rate from 4.2\% to 2.25\% using only 31\% network parameters of the baseline CNN. Combined with TIPooling, ORN further decreases the error rate to 1.54\%, achieving state-of-the-art performance, which shows that ORNs have good generalization capability for such reduced training sample cases.

\subsection{Orientation Estimation}
\label{sec:MNIST-estimate}
    ORN is evaluated on the weakly image orientation estimation problem, using the STN \cite{Jaderberg2015} as the baseline. The training images have only class labels but lack orientation annotation, which is estimated during learning. We upgrade the localisation sub-network of STN from a conventional CNN to ORN by converting Conv layers to ORConv layers which use ARFs with eight orientation channels. The STN model is simplified to process rotation only, which means that its localisation network estimates only a rotation parameter.

    \begin{table}
        \begin{center}
            \begin{tabular}{|l|c|c|}
                \hline
                Method              & Std & Error(\%)       \\ \hline\hline
                STN \cite{Jaderberg2015}  & 0.745    & 3.38 \\
                OR-STN(ORAlign)     & 0.749 & 3.61          \\
                OR-STN        & \textbf{0.397} & \textbf{2.43} \\ \hline
            \end{tabular}
        \end{center}
        \caption{
            Orientation estimation performance. The second column describes the standard deviation of calibrated orientations and the third column describes the classification error rates.
        }
    \label{tab:MNIST-orientation-estimate}
    \vspace{-0.2em}
    \end{table}

    \begin{figure}
        \centering
        \includegraphics[width=1\linewidth]{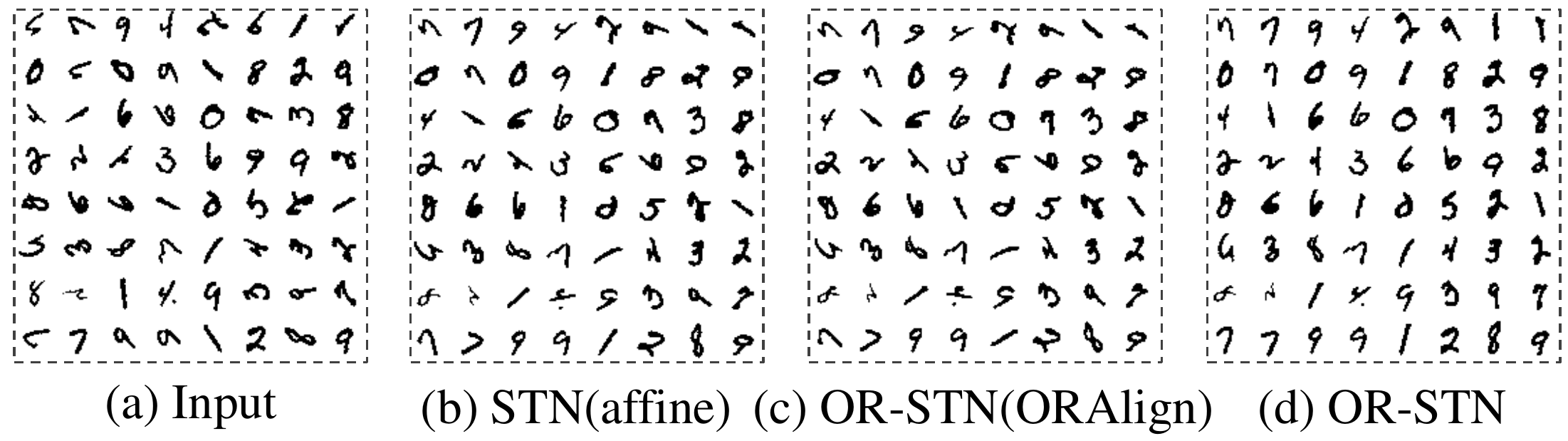}
        \caption{Orientation estimation. (a) is a mini-batch of samples from \textit{MNIST-half-rot} and (b)-(d) are their rotation-rectified results.}
    \label{fig:Estimation}
    \vspace{-1em}
    \end{figure}

    \begin{figure*}[t]
        \centering
        \subfigure[STN] {
            \label{fig:Distribution:a}
            \includegraphics[width=0.32\linewidth, height=0.17\linewidth]{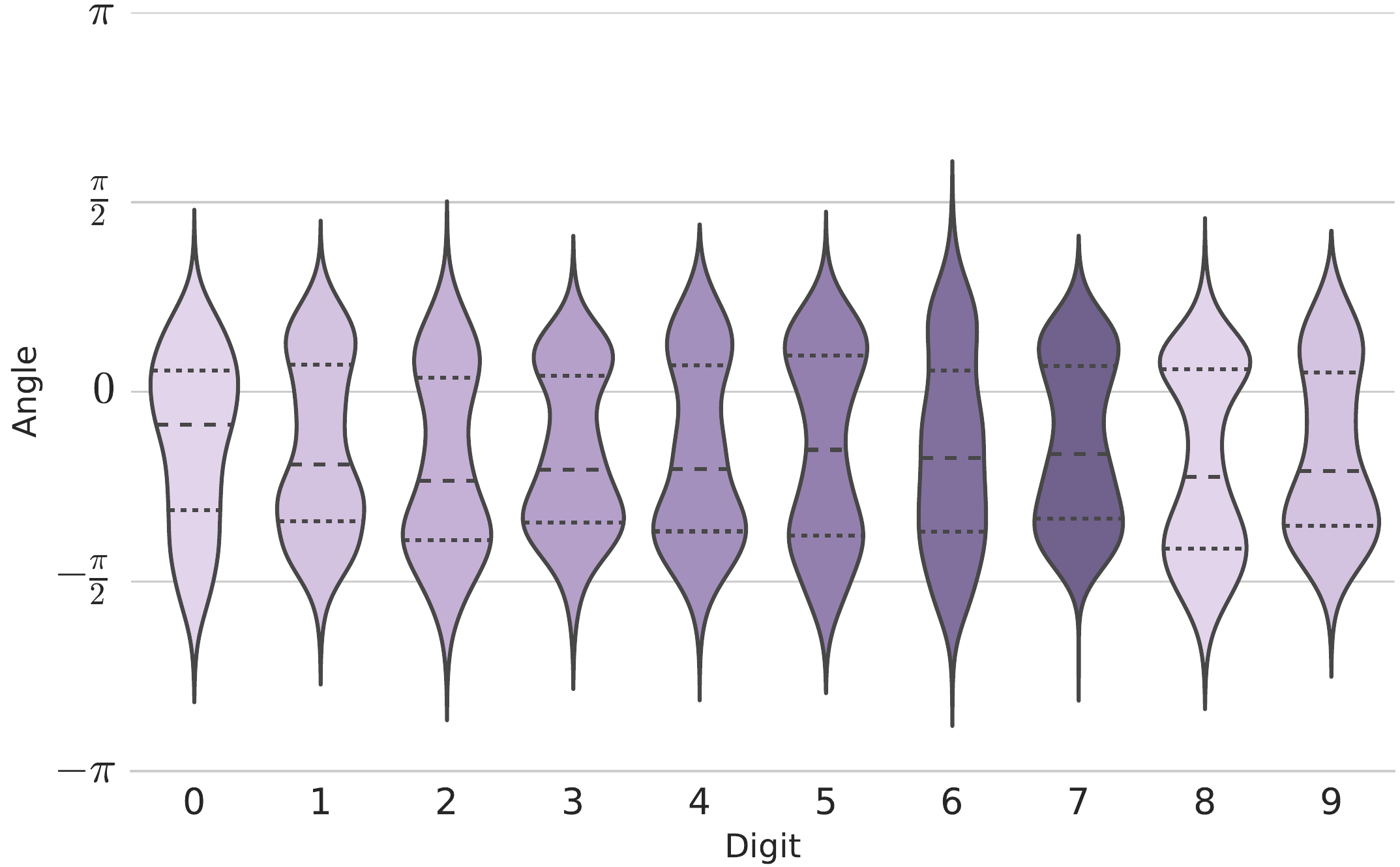}
        }
        \subfigure[OR-STN(ORAlign)] {
            \label{fig:Distribution:b}
            \includegraphics[width=0.32\linewidth, height=0.17\linewidth]{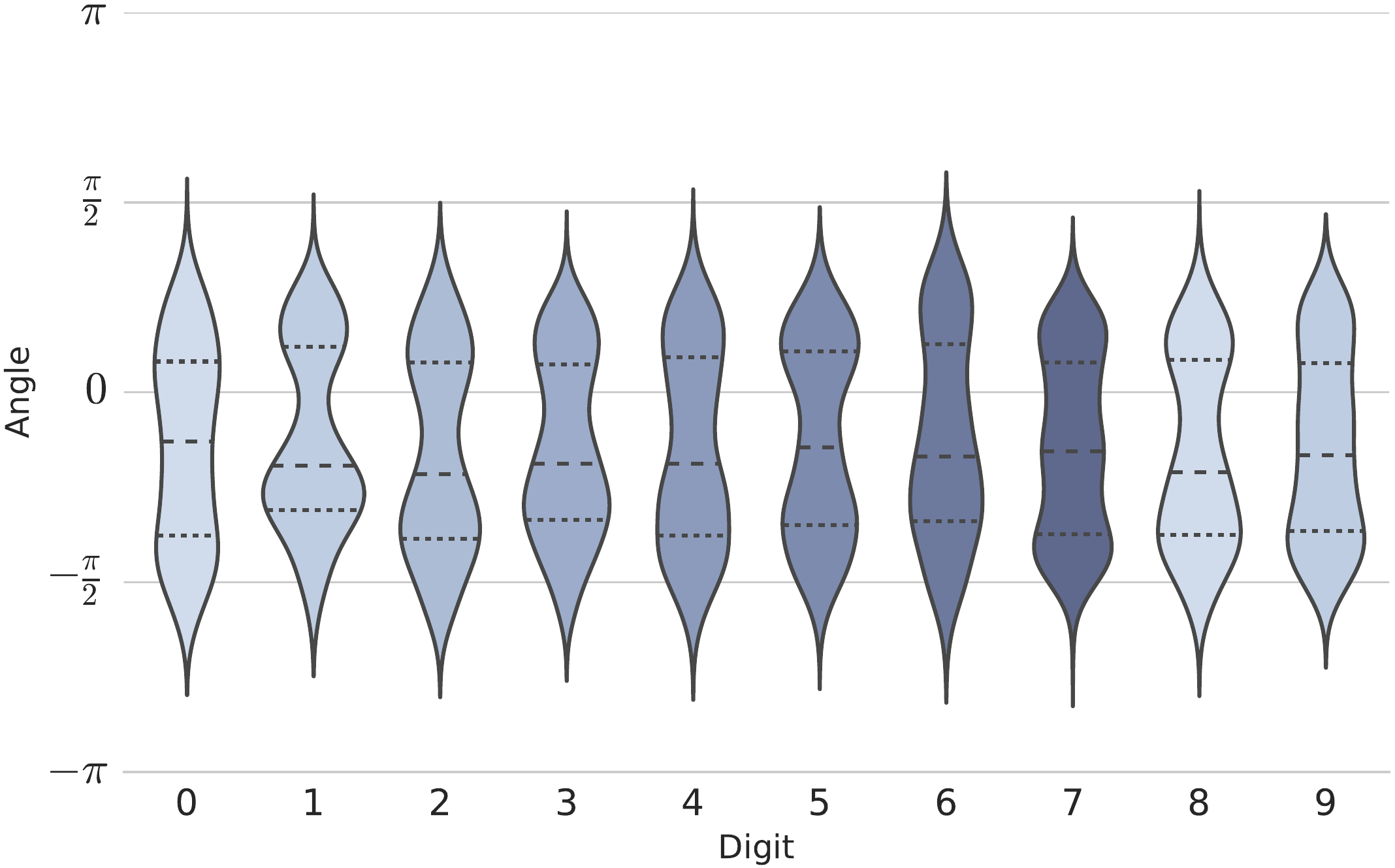}
        }
        \subfigure[OR-STN] {
            \label{fig:Distribution:c}
            \includegraphics[width=0.32\linewidth, height=0.17\linewidth]{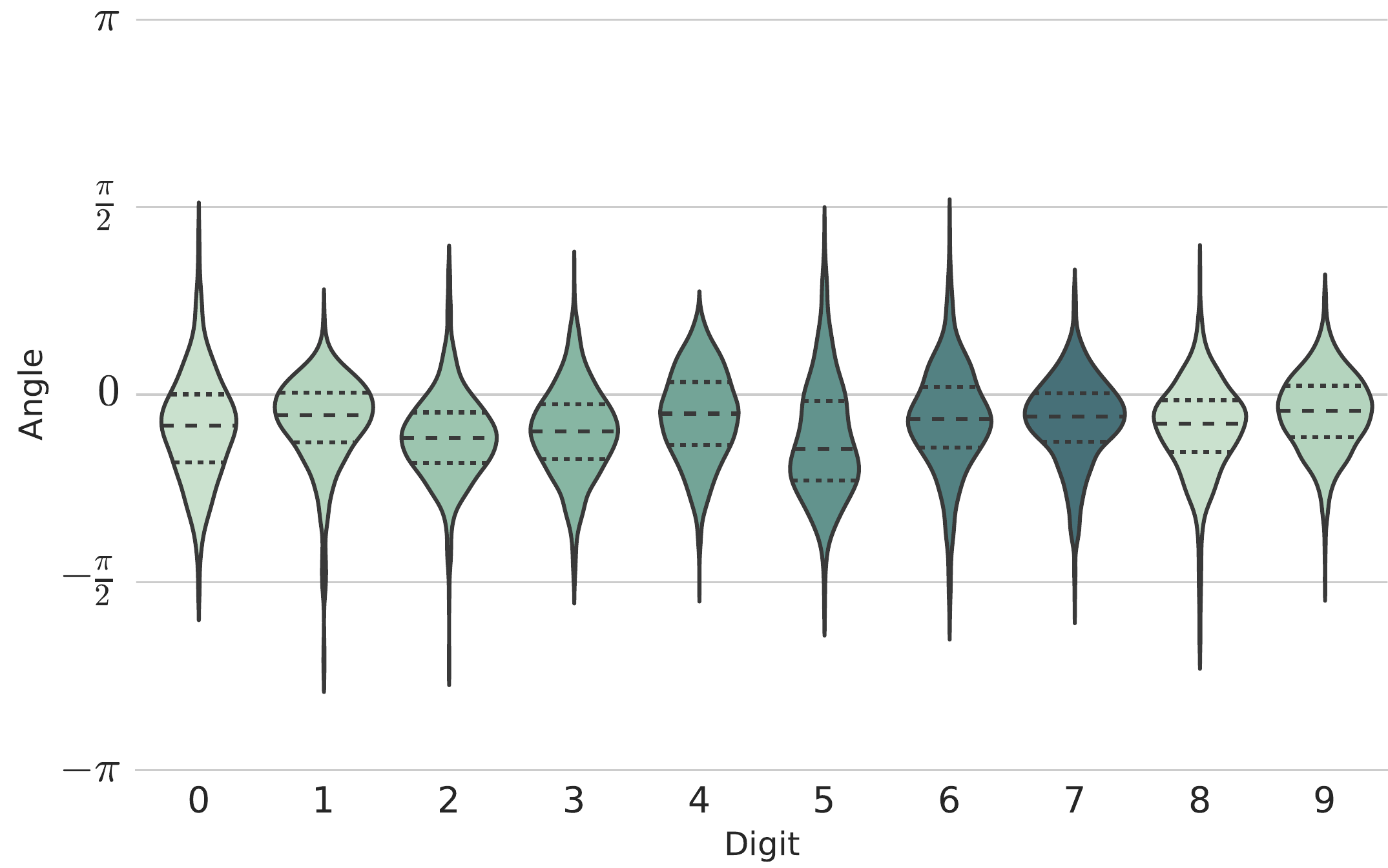}
        }
        \caption{Distributions of samples' orientations after rotation-rectification.}
    \label{fig:Distribution}
    \vspace{-0.4em}
    \end{figure*}

    STN, OR-STN and OR-STN(ORAlign) are trained on the \textit{MNIST-half-rot} dataset which is built by randomly rotating each sample in the MNIST dataset in the range $[-\frac{\pi}{2}, \frac{\pi}{2}]$ (half the circle). All the networks use hyper-parameters as Sec.~\ref{sec:MNIST-rot} and are trained by only 80 epochs to make the localisation sub-network converge. The orientation estimation results are presented in Tab.~\ref{tab:MNIST-orientation-estimate}, the rotation-rectified images are shown in Fig.~\ref{fig:Estimation}, and angle statistics of rotation-rectified images are shown in Fig.~\ref{fig:Distribution}. It can be seen in Fig.~\hyperref[fig:Estimation]{\ref*{fig:Estimation}(b)} that STN cannot effectively handle the large-angle rotation problem, because the localisation sub-network itself is a conventional CNN, lacking the ability to explicitly process significant rotation. When upgrading the localisation network of STN to ORN (without ORAlign), it can be seen in Fig.~\hyperref[fig:Estimation]{\ref*{fig:Estimation}(d)} that most digit orientations are correctly estimated. In Fig.~\ref{fig:Distribution:b}, it can be seen that the OR-STN(ORAlign) performs even worse than the baseline on orientation estimation, because after the feature alignment, features become rotation-invariant and thus lose orientation information.
    Tab.~\ref{tab:MNIST-orientation-estimate} shows that upgrading localisation sub-network to ORN significantly improves the performance. Such experiments validate that the ARFs can capture the orientation information of discriminative patterns and explicitly encode them into feature maps with orientation channels, which are effective for image orientation estimation.

\subsection{Natural Image {Classification}}
    \begin{table}
        \begin{center}
        \footnotesize
        \setlength{\tabcolsep}{4pt}
        \begin{tabular}{|l|cccc|}
                \hline
                Method                           & depth-k & params & CIFAR10(\%)  & CIFAR100(\%)  \\ \hline\hline
                NIN \cite{Lin2014}                           & -       & -      & 8.81          & 35.67          \\
                DSN \cite{Lee2015}                           & -       & -      & 8.22          & 34.57          \\
                Highway \cite{Srivastava2015}                       & -       & -      & 7.72          & 32.39          \\
                ELU \cite{Clevert2015}                           & -       & -      & 6.55          & 24.28          \\ \hline
                VGG \cite{Simonyan2014}                           & 16      & 20.1M  & 6.32          & 28.49          \\ \hline
                \textbf{OR-VGG}                  & 16-$\tfrac{1}{2}$  & 10.1M  & 5.47          & 27.03          \\ \hline
                ResNet \cite{He2015}       & 110     & 1.7M   & 6.43          & 25.16          \\ \hline
                \textbf{OR-ResNet}               & 110-$\tfrac{1}{2}$ & 0.9M   & 5.31          & -              \\ \hline
                \multirow{3}{*}{pre-act-ResNet\cite{He2016}}   & 110     & 1.1M   & 6.37          & -              \\
                                                 & 164     & 1.7M      & 5.46          & 24.33          \\
                                                 & 1001    & 10.3M  & 4.92          & 22.71          \\ \hline
                \multirow{3}{*}{WideResNet\cite{Zagoruyko2016}}      & 40-4    & 8.9M   & 4.53          & 21.18          \\
                                                 & 16-8    & 11.0M  & 4.27          & 20.43          \\
                                                 & 28-10   & 36.5M  & 3.89          & 18.85          \\ \hline
                \multirow{4}{*}{\textbf{OR-WideResNet}}  & 40-$\tfrac{1}{2}$    & 1.1M   & 4.34          & 23.19              \\
                                                 & 40-2    & 4.5M   & 3.43          & 18.82          \\
                                                 & 28-5    & 18.2M  & \textbf{2.98} & \textbf{16.15} \\ \hline
        \end{tabular}
        \end{center}
        \caption{
            Results on the natural image classification benchmark. In the second column, $k$ is the widening factor corresponding to the number of filters in each layer.
        }
    \label{tab:CIFAR}
    \vspace{-1em}
    \end{table}

\label{sec:CIFAR}
    Although most objects in natural scene images are upright, rotations could exist in small and/or medium scales (from edges to object parts). It is interesting to validate whether ORNs are effective to handle such partial object rotation or not.
    CIFAR-10 and CIFAR-100 datasets \cite{Krizhevsky2009} consist of 32x32 real-world object images drawn from 10 and 100 classes split into 50,000 training and 10,000 testing images. Three DCNNs including VGG \cite{Simonyan2014}, ResNet \cite{He2015} and WideResNet \cite{Zagoruyko2016}, are used as baselines on these datasets. Promoting the baselines to ORNs is done by converting each Conv layer to an ORConv layer that uses ARFs with eight orientation channels, and using an additional ORAlign layer to encode rotation invariant representations.

    Following the \textbf{V2} settings of WideResNet \cite{Zagoruyko2016}, image classification results, Tab.~\ref{tab:CIFAR}, show that ORNs consistently improved baselines with much fewer parameters.
    For example, OR-VGG uses about 50\% parameters of the baseline to achieve better results. OR-WideResNet-40-2 (without dropout) uses only 12\% parameters (4.5M vs 36.5M) to outperform the state-of-the-art WideResNet-28-10 (with dropout) on CIFAR10. OR-WideResNet-28-5 (with dropout) uses about 50\% parameters of the baselines yet significantly improve the state-of-the-arts on both CIFAR10 and CIFAR100.
    \begin{figure}
        \begin{center}
            \includegraphics[width=0.9\linewidth]{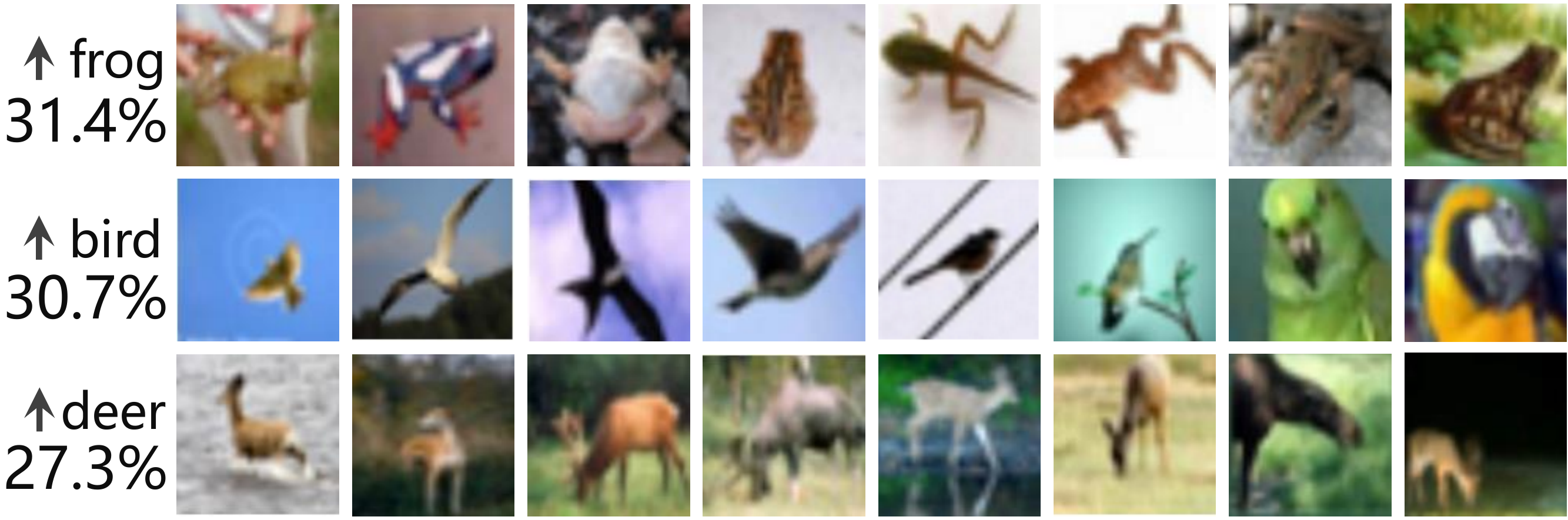}
        \end{center}
        \caption{Sample images that contain rotated objects/parts falsely classified by the ResNet but correctly recognized by the proposed ORNs in CIFAR10.}
    \label{fig:CIFAR-Vis}
    \vspace{-0.5em}
    \end{figure}
    The top-3 improved classes of CIFAR10 are \textbf{frog} (31\% higher than baseline ResNet), \textbf{bird} (30.7\%) and \textbf{deer} (27.3\%), which happen to involve significant local and/or global rotations, Fig.~\ref{fig:CIFAR-Vis}. This further demonstrates the capability of ORN to process local and global image rotations.
\section{Conclusions}
    In this paper, we proposed a simple but effective strategy to explicitly encode hierarchical orientation information of discriminative patterns and handle the global/local rotation problem. The primary contribution is designing Active Rotating Filters (ARFs), as well as upgrading the state-of-the-art DCNN architectures, \eg, VGG, ResNet, STN, and TI-Pooling, to Oriented Response Networks (ORNs). Experimentally, ORNs outperform the baseline DCNNs while using significantly fewer (12\%-50\%) network parameters, which indicates that the usage of model-level rotation prior is a key factor in training compact and effective deep networks.

\section*{Acknowledgements}
    The authors are very grateful for support by NSFC grant 61671427, BMSTC grant Z161100001616005, STIFCAS grant CXJJ-16Q218, and NSF.

{\small
\bibliographystyle{ieee}
\bibliography{egbib}
}

\end{document}